\documentclass[lettersize,journal]{IEEEtran}
\usepackage{color}
\usepackage{CJKutf8}
\usepackage{amsmath,amsfonts}
\usepackage{algorithmic}
\usepackage{algorithm}
\usepackage{array}
\usepackage[caption=false,font=normalsize,labelfont=sf,textfont=sf]{subfig}
\usepackage{textcomp}
\usepackage{stfloats}
\usepackage{url}
\usepackage{verbatim}
\usepackage{graphicx}
\usepackage{nicematrix}
\usepackage{cite}
\usepackage{booktabs}
\usepackage{multirow}
\usepackage{enumitem}

\graphicspath{{figs/}}
\begin{document}
	\begin{CJK*}{UTF8}{gbsn}	
		\title{Communication-Efficient Federated Learning with Adaptive Compression
			under Dynamic Bandwidth}  
		
		\author{Ying Zhuansun, Dandan Li, Xiaohong Huang, and Caijun Sun
			\thanks{This work was supported by the National Key Research and Development Program of China under Grant No.2020YFE0200500. (Corresponding Author: Dandan Li)}
			\thanks{Ying Zhuansun, Dandan Li, and Xiaohong Huang are with the School of
				Computer Science (National Pilot Software Engineering School), Beijing
				University of Posts and Telecommunications, Beijing 100876, China (e-mail:  zhuansunying@bupt.edu.cn; dandl@bupt.edu.cn; huangxh@bupt.edu.cn).}
			\thanks{Caijun Sun is with Zhejiang Lab, Hangzhou, China, (e-mail:
				sun.cj@zhejianglab.edu.cn).}
			}
			
		
		
		\maketitle
		
		\begin{abstract}
			Federated learning can train models without directly providing local data to the server. However, the frequent updating of the local model brings the problem of large communication overhead. Recently, scholars have achieved the communication efficiency of federated learning  mainly by model compression. But they ignore two problems: 1) network state of each client changes dynamically; 2) network state among clients is not the same.
			The clients with poor bandwidth update local model slowly, which leads to low efficiency. 
			To address this challenge, we propose a communication-efficient federated learning algorithm with adaptive compression under dynamic bandwidth (called AdapComFL). 
			Concretely, each client performs bandwidth awareness and bandwidth prediction.
			Then, each client adaptively compresses its local model via the improved sketch mechanism based on his predicted bandwidth. Further, the server aggregates sketched models with different sizes received. 
			To verify the effectiveness of the proposed method, the experiments are based on real bandwidth data which are collected from the network topology we build, and benchmark datasets which are obtained from open repositories. 
			We show the performance of AdapComFL algorithm, and compare it with existing algorithms. 
			The experimental results show that our AdapComFL achieves more efficient communication as well as competitive accuracy compared to existing algorithms.   
		\end{abstract}
		
		\begin{IEEEkeywords}
			Federated learning, communication efficiency, dynamic bandwidth, sketch.
		\end{IEEEkeywords}
		
		\section{Introduction}
		\IEEEPARstart{T}{he} development of communication technologies catalyzes  diverse Internet of Things (IoT) scenarios, then the massive volumes of data generated by IoT clients can be fully employed to make people's life more convenient, such as smart healthcare \cite{mcmahan2017communication,zhang2017security,petrolo2017towards}. Generally, clients need transport their own data to third party for training, which exposes them to the risk of privacy leakage~\cite{voigt2017eu}. 
		Fortunately, federated learning, which allows clients to participate in distributed training without move their raw data~\cite{kairouz2021advances,yang2019federated,li2020federated}, bridges a light for machine learning with privacy protection. 
		Compared with the framework of central machine learning, clients and the server require more rounds of communication in the framework of federated learning, which brings the communication overhead. In fact, communication cost becomes bottleneck of practical federated learning.

		
		\par To achieve the communication efficiency of federated learning, there are two categories of approaches: 
		(1) reducing the frequency of total communication by increasing the amount of local	computation \cite{mcmahan2017communication}, \mbox{Kone{\v{c}}n{\`y} \emph{et al}. \cite{konevcny2017stochastic}} allow more local updates per communication round, \mbox{Li \emph{et al}. \cite{li2020federatedoptimization}} dynamic adjusts the local calculation epoch of each client. These methods require powerful computing resources, and due to the emphasis on local training, the global model may have performance differences among clients. 
		(2) reducing the volume of messages each round of communication. It uses different methods to compress local models, including stochastic sparsification~\cite{konevcny2016federated,lin2017deep,ji2021dynamic,sattler2019robust,li2021communication}, quantization~\cite{alistarh2017qsgd,amiri2020federated,reisizadeh2020fedpaq,konevcny2016federated,xu2020ternary,chang2020communication,jhunjhunwala2021adaptive,he2020cossgd} and pruning ~\cite{zhang2022fedduap,jiang2022model}.
		
		
		Stochastic sparsification mainly uses top-k sparsification method which filters the more important model parameters \cite{lin2017deep,ji2021dynamic}.		
		For quantization including probability quantization	and gradient quantization methods, the probability quantization characterizes the model parameters as extremum~\cite{konevcny2016federated}.
		Gradient quantization considers model precision, which assigns different precisions according to the importance of the model parameter~\cite{xu2020ternary,chang2020communication,jhunjhunwala2021adaptive}. Moreover, some works combine quantization with stochastic sparsification~\cite{konevcny2016federated,sattler2019robust,li2021communication}.
		For pruning, there are two main categories: structured pruning and unstructured pruning. Structured pruning reduces the model size by cutting out the unimportant parts of the model structure~\cite{zhang2022fedduap}, while unstructured pruning maintains the original model size but cuts out unimportant model parameters~\cite{jiang2022model}.
		The works mentioned above just focus on communication efficiency in federated learning but ignore privacy-preserving issues. \mbox{Liu \emph{et al}. \cite{liu2019enhancing}} trade-off communication efficiency and privacy via the sketch compression method.
		
		\par Remarkably, existing methods of achieving communication efficiency in federated learning ignore two problems. First, the network state of each client changes dynamically, as shown by changes in bandwidth. Second, bandwidth of clients is different. For clients with poor bandwidth, the server needs to delay the aggregation of the global model because these clients upload the local model slowly. 
		Here, we address this challenge and propose a communication-efficient federated learning algorithm with adaptive compression under dynamic bandwidth, which is called AdapComFL. The workflow of AdapComFL algorithm is as follows in brief.
		(1) Clients carry out bandwidth awareness during local model training. 
		(2) After training, clients predict bandwidth, and compress the model gradient adaptively by the improved sketch mechanism based on the predicted bandwidth.
		(3) Clients send the local sketch model with different sizes to the server for aggregation. Note that, these sketch models have the same columns but rows differ. After aligning the sizes and linearly accumulating all the sketch models, the row-wise average is calculated based on the accumulated times for each row. 
		(4) Clients receive aggregated sketch model and decompress it.

		\par The contributions of this article are summarized as follows:
		
		\begin{itemize}
			\item{Firstly, to solve the problem of communication efficiency in federated learning, which ignores the dynamic and difference of clients' bandwidth, we propose a communication efficiency federated learning algorithm with adaptive compression under dynamic bandwidth.}
			\item{Secondly, to compress the model adaptively, we improved the sketch mechanism. We elastically adjust the rows of the sketch model to change the size adaptively according to the predicted bandwidth. We then address the accuracy issue resulting from the sketch mechanism. In the compression process, we filter the model gradients mapped to the same position based on the coefficient of variation~\cite{abdi2010coefficient}, rather than adding them together. Furthermore, to aggregate sketch models of different sizes, we perform size alignment, accumulate the sketch models, and calculate the row-wise averages. }
			\item{Finally, to confirm that our framework can achieve communication efficiency in federated learning, we build a network topology to collect real bandwidth data. We then conduct comparative experiments with other algorithms.}
		\end{itemize}
		
		\par The rest of organize in this article as follows: Section II briefly introduces some model compression methods in federated learning. Section III proposes a method to reduce communication overhead with adaptive compression under dynamic bandwidth. The setups and results are presented in Section IV to perform experiments and demonstrate the effectiveness of AdapComFL. Section V concludes the article based on our contributions.

		\section{Related Work}
		
		To reduce federated learning communication overhead, the typical methods include increasing local computation and model compression.  
		In this section, we review some methods	for model compression from stochastic sparsification, quantization, distillation, pruning, and sketch.
		\subsection{Stochastic Sparsification-based Compression Method}
		Stochastic sparsity achieves communication efficiency optimization with the method of describing the local model by a sparse matrix through a predefined random sparse pattern \cite{konevcny2016federated}. 
		\mbox{Lin \emph{et al}. \cite{lin2017deep}} proposed an algorithm, which uploads the model gradient when its value exceeds a certain threshold to achieve communication efficiency. In \cite{ji2021dynamic}, a top-$k$ selection algorithm was proposed to improve communication efficiency. Each client examines the parameter differences between the current model and the global model and picks the top-$k$ greatest-difference parameters to upload. \mbox{Sattler \emph{et al}. \cite{sattler2019robust}} proposed a novel Sparse Ternary Compression (STC) framework, which extends the existing top-$k$ gradient sparse compression technique via ternary, error accumulation, and optimal Golomb encoding. \mbox{Han \emph{et al}. \cite{han2020adaptive}} proposed adaptive gradient sparsification, which set different sparse parameters for different rounds. \mbox{Ozfatura \emph{et al}. \cite{ozfatura2021time}} sought sparse correlation at consecutive iterations in FL to obtain the positions of important values, and simply upload the values at these positions.
		
		\subsection{Quantization-based Compression Method}
		About compression schemes in quantization, including probability quantization and gradient quantization. In probability quantization, the model weight is compressed to extremum~\cite{konevcny2016federated}. 
		For gradient quantization, the model reduces the precision.  \mbox{Xu \emph{et al}. \cite{xu2020ternary}} proposed the Federated Training Trivial Quantization (FTTQ) method to dynamically quantize the model parameters. In addition to model parameters, the quantization factor is uploaded, aggregated, and updated. Thereby, FTTQ dynamic quantifies the model parameters by dynamic factor. \mbox{Chang \emph{et al}. \cite{chang2020communication}} applied Multiple Access Channel (MAC) technologies to propose a MAC perceptual gradient quantization scheme. They performed quantization via gradient informativeness and channel condition to utilize the communication resources. Different from previous works, \mbox{He \emph{et al}. \cite{he2020cossgd}} proposed a nonlinear quantization scheme based on cosine function. They divide the distribution space of the model nonuniformly by finer quantization intervals. \mbox{Jhunjhunwala \emph{et al}. \cite{jhunjhunwala2021adaptive}} considered the error, adjusted the quantization level automatically.
		\subsection{Pruning-based and Distillation-based Compression Method}
		About compression schemes in pruning, including structured pruning, and unstructured pruning. Specifically, structured pruning removes weights with neurons, filters, or channels. \mbox{Zhang \emph{et al}. \cite{zhang2022fedduap}} proposed the Federated Adaptive Structured Pruning (FedAP). It prunes filters according to the non-Independent and Identically Distributed (non-IID) degree of local data, which reduces communication costs while ensuring accuracy. Unstructured pruning removes unimportant weights in the model. For example, in \cite{jiang2022model}, before the federal learning, the server first chooses a client to prune the initial model. Then, the global model is continuously pruned in the aggregation of federated learning.
		\par For distillation, \mbox{Jeong \emph{et al}. \cite{jeong2018communication}} proposed a federated distillation strategy that combined knowledge distillation with federated learning. They achieve communication overhead savings by exchanging model output instead of their parameters. \mbox{Sattler \emph{et al}.} \cite{sattler2020communication} proposed Compressed Federated Distillation (CFD), which achieves communication efficiency by analyzing the effects of active distillation-data curation, soft-label quantization, and delta-coding techniques.
		\subsection{Sketch-based Compression Method}
		The sketch characterizes the model by several independent hash functions. In 2019,  \mbox{Li \emph{et al}. \cite{li2019privacy}} applied the sketch algorithm from statistics~\cite{charikar2002finding} to distributed scenario by improving it. Then \mbox{Liu \emph{et al}. \cite{liu2019enhancing}} took the lead in proposing a sketch-based optimization algorithm for federated learning communication efficiency.  Before clients upload the local model, they use the sketch method for compression, then the server aggregates and broadcasts the sketch model. Finally, clients decompress the sketch model and train. Thereby, they balance communication efficiency and privacy. Moreover, \mbox{Rothchild \emph{et al}. \cite{rothchild2020fetchsgd}} combined sketch with error momentum to improve accuracy. The momentum sketch model and error sketch model were cached in model aggregation process. They made up for the errors caused by sketch mechanism. However, they decompress sketch model on the server, which reduce security.
		
		\par Unfortunately, federated learning is still challenged by network bandwidth. Different from above works, we propose a federated learning algorithm for efficiency and security, which adaptively compress under dynamic bandwidth.

		\section{Communication-Efficient Federated Learning with Adaptive Compression under Dynamic Bandwidth}
		In this section, to address the dynamic of each client and the differences among clients in network bandwidth, we construct the model and design a communication-efficient federated learning algorithm.
		
		\subsection{Model Construction}
		In traditional federated learning scenario, without loss of generality, we take the $r$-th round as an example, suppose there are $C$ clients, $1 \le i \le C$, client $i$ trains model and uploads it to server. Then the server aggregates models and sends the aggregated model to each client. Repeat the above steps until global model converges. The optimization objective is: 
				
		\begin{equation}
			\setlength\abovedisplayskip{-10pt}
			\begin{aligned} 
			    {\min}  \quad F(w) = \sum\limits_{i=1}^{C} {\frac{N_{i}}{N}{F_{i}}({w_{i}})} 
			\end{aligned},
			\label{optimizationfl}
		\end{equation}
		where $F$ is global loss function, $N_{i}$ is the number of samples for client $i$, $N$ is the total number of samples, $w_{i}$ is local model weights. 
		\par
\begin{figure}[t]
			\centering
			\includegraphics[scale=0.54]{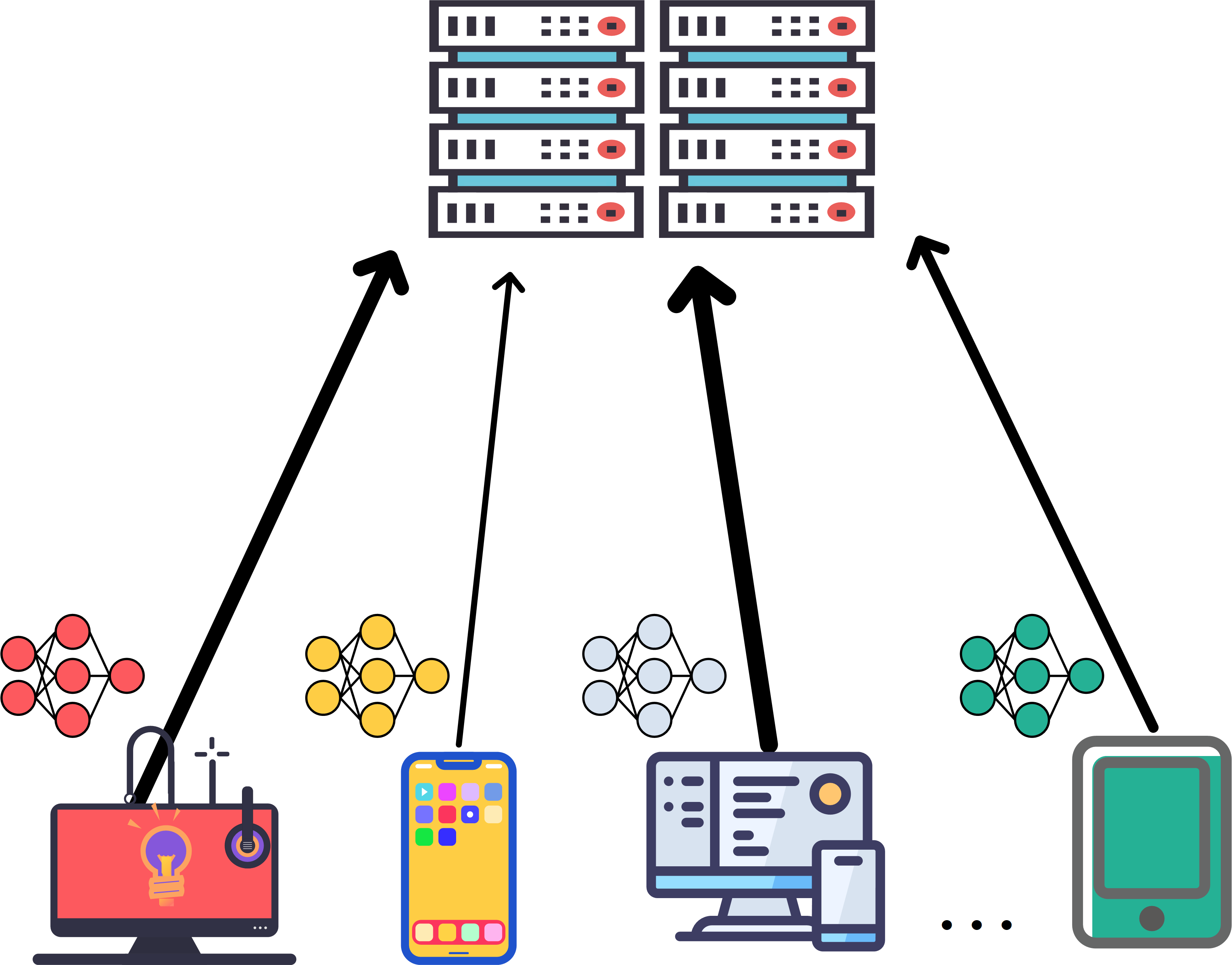}
			\caption{The federated Learning under dynamic bandwidth. The thickness of the line represents the condition of the bandwidth.
			}
			\label{fig:overview}
		\end{figure}

 In fact, the models that clients upload have large data volumes, but the bandwidths among clients are different, and the bandwidth of each client is dynamic. The overview is illustrated in Fig.~\ref{fig:overview}. Clients with poor bandwidth are waited for by the server for aggregate models, leading to low efficiency. 
		Here, we consider the above bandwidth issues in federated learning scenario, propose a communication-efficient federated learning algorithm with adaptive compression under dynamic bandwidth, AdapComFL. 
		\par Different from traditional federated learning, client $i$ collects bandwidth data $\pmb{B}_{i}$ while training, predicts bandwidth $b_{i}$ of uplink communication, obtains data volume ${D_{i}}$ of uplink communication, subsequently determines the data volume ${D'_{i}}$ of the upload model, and compresses model gradient by compression operator $\mathcal{C(\cdot)}$. The optimization objectives of AdapComFL are as follows:
				
		\begin{equation}
			\begin{cases}

			\begin{aligned} 
				& \min  \quad F(w) = \sum\limits_{i=1}^{C} {\frac{{{N_{i}}}}{N}{F_{i}}({w_{i}}, {w_{b,i}},T'_{i}, \mathcal{C}, \mathcal{D})} & \\
				& \min_{1\le i\le C} \quad T'_{i} = {\frac{{D'_{i}}}{b'_{i}{log}_2 (1+{S\!N\!R})}}
			\end{aligned},
			\end{cases}\label{optimization}
		\end{equation}			
		where $w_{b,i}$ is local model weights for bandwidth prediction, $\mathcal{D(\cdot)}$ is decompression operator, $T'_{i}$ is the real time consumed for uplink communication, $b'_{i}$ is real bandwidth, ${S\!N\!R}$ is the signal-to-noise ratio (SNR). 

		\begin{figure*}[htbp]
			
			\centering
			\includegraphics[scale=0.37]{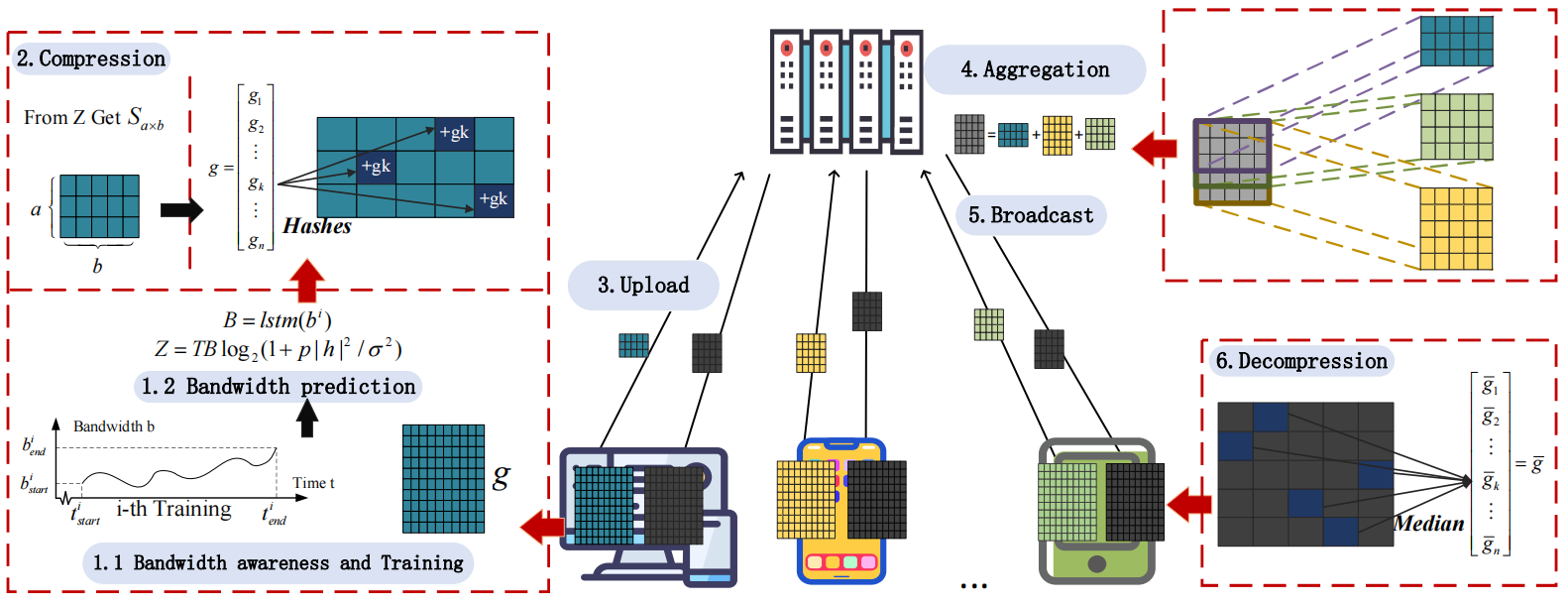}
			
			\caption{AdapComFL. Firstly, each client carries out bandwidth awareness while training the model, and predicts bandwidth based on aware data to obtain upload data volume, as in Step 1. Thus, the gradient is adaptively compressed and uploaded to the server, i.e., Steps 2-3. Then, the server aggregates all sketch models of different sizes into one and sends it, i.e., Steps 4-5. Finally, the client decompresses the updated sketch model as presented in Step 6.}
			\label{fig:sketch2}
		\end{figure*}

		\subsection{Algorithm}
		In this section, we propose the AdapComFL algorithm, which the workflow is illustrated in Fig.~\ref{fig:sketch2}.
		\par 1) \emph{Bandwidth Awareness and Prediction}: Each client is aware of the bandwidth during local model training, predicts bandwidth based on collected data, obtains the data volume of uplink communication, and determines the size of sketch model.
		\par 2) \emph{Compression}: Each client adaptively compresses local model to sketch model. 
		\par 3) \emph{Upload}: Each client uploads the sketch model to server.
		\par 4) \emph{Aggregation}: Server aggregates the sketch models which have different sizes.
		\par 5) \emph{Broadcast}: Server sends aggregated sketch to each client.
		\par 6) \emph{Decompression}: Each client decompresses sketch to recover the model.
		\par Repeat steps 1-6 until the model converges. The detail of AdapComFL is shown in Algorithm~\ref{alg1}.
		
		In AdapComFL, to address the issues of communication bottleneck and improve accuracy in federated learning, there are two differences from traditional federated learning: bandwidth awareness and prediction; dynamic bandwidth-based local model compression. 
		
		\begin{algorithm}[t]
			\caption{Federated Learning with Adaptive Bandwidth Compression under Dynamic Bandwidth (AdapComFL)}
			\begin{algorithmic}[1]
				\REQUIRE The number of communication rounds $R$, number of Clients $C$, communication time $\text{T}$, SNR $SNR$, hash functions $h$, compression operator $\mathcal{C(\cdot)}$, decompress operator $\mathcal{D(\cdot)}$, aggregation operator $\mathcal{AGG(\cdot)}$
				\STATE Initialize $w^0$ on the clients
				\STATE Initialize $S^0$ to zero sketch model
				\FOR{$r \in R$}
				\FOR{$i \in C$}
				\IF{$r \neq 0$}
				\STATE Updating model:$w_{i,r}=w_{i,r-1}+\mathcal{D}(S_{agg,r-1})$
				\ENDIF
				\STATE Start collecting bandwidth data $\pmb{B}_{i,r}$
				\STATE Start local training $g_{i,r}=\eta \nabla F\left(w_{i,r-1}\right)$
				\STATE Predict bandwidth $b_{i,r}$ based $\pmb{B}_{i,r}$ and stop collect data
				\STATE Obtain the data volume of communication:
				$D_{i,r} =\text{T}b_{i,r} \log _2\left(1+SNR\right)$
				\STATE Compression according to $D_{i,r}$: $S_{i,r} ={\mathcal{S}}\left(g_{i,r}\right)$
				\STATE  Send $S_{i,r}$ to the server
				\ENDFOR
				\STATE {Aggregate sketch models $S_{{agg,r}}\!=\!\mathcal{AGG} \left(\left\{S_{i,r}, 1 \leq i \le C\right\}\right)$}
				\ENDFOR
				
				\ENSURE $S$\\ 
			\end{algorithmic}
			\label{alg1}
		\end{algorithm}
		
		\par {\bf{Bandwidth awareness and prediction: }}In order to achieve communication efficiency in federated learning, each client first predicts the network bandwidth, computes the data volume of uplink communication, acquires the size of sketch model, and then, adaptively compresses the local model to sketch model.
		\par In order to predict bandwidth, client $i$ first continuously collects bandwidth data $\pmb{B}_i\in R^m$ via Iperf~\cite{tirumala1999iperf} while training, where $\pmb{B}_i$ is:
		\begin{equation}
			\label{Bi}
			\pmb{B}_i=[B_{i,1},B_{i,2},B_{i,3},\dots,B_{i,m}].
		\end{equation}
		The bandwidth is predicted via a Long Short Term Memory (LSTM) recurrent neural network with one input layer, one output layer, and two hidden layers~\cite{mei2020realtime}. 
		Suppose the sequence length of LSTM is $f$, $f<m$, the dataset is $\pmb{B}_{i}$, during $j-th$ batch, the input data are $[B_{i,j},B_{i,j+1},\dots,B_{i,j+f-1} ]$, the label data is $B_{i,j+f}$. 
		When the training is completed, the output of LSTM is the predicted bandwidth $b_i$. 
		\par Therefore, the data volume of uplink communication $D_{i}$ is obtained~\cite{xu2020client}:
		
		\begin{equation}
			\label{deqn_ex6}
			D_{i}=\text{T}{b}{_{i}}{log}_2 (1+{S\!N\!R}).
		\end{equation}

		\par {\bf{Dynamic bandwidth-based local model compression in federated learning: }}To adaptively compress the local model gradient to the size of $D_{i}$, we improve the sketch mechanism.
		
		\begin{figure*}[htbp]
			
			\centering
			\includegraphics[scale=0.55]{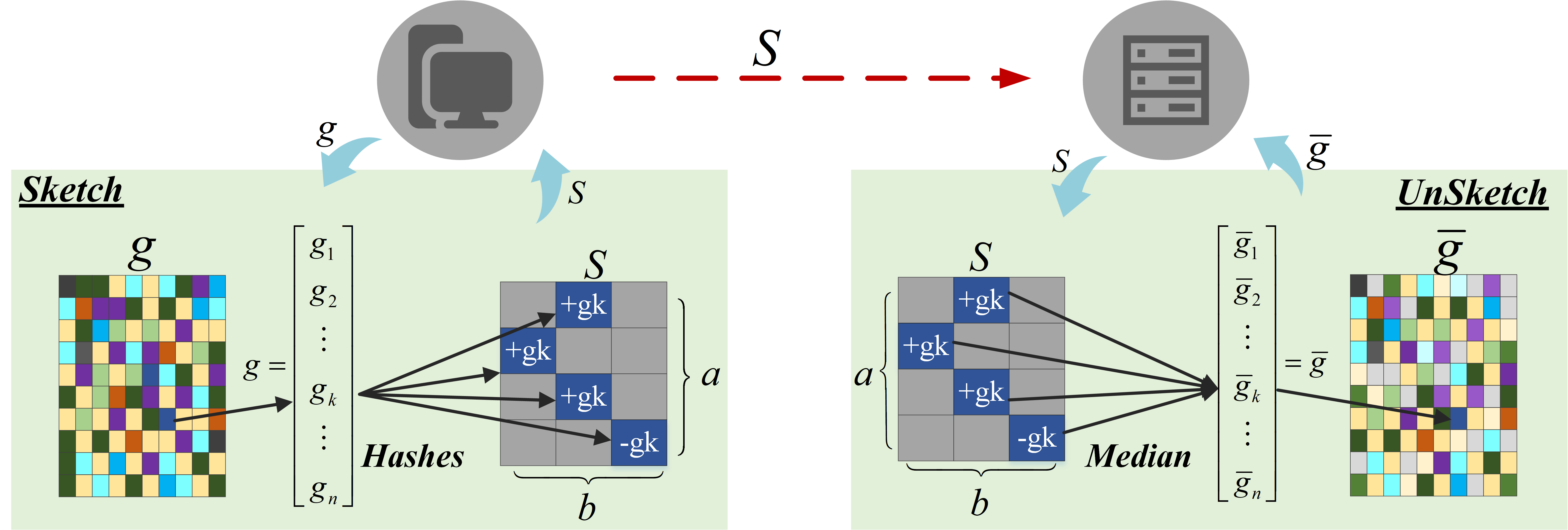}
			\caption{Compress and decompress in the sketch mechanism. The left node is the client and the right node is the server. The left node compresses the model gradient $\pmb{g}$ to sketch model $S$ and transmits it to the right node. The right node decompresses $S$ to obtain $\overline{\pmb{g}}$.}
			
			\label{fig:sketch1}
		\end{figure*}
		
		Generally, before client $i$ uploads the model gradient vector $\pmb{g}_i\in$$R^{n}$, the $\pmb{g}_i$ is compressed into sketch model with an $a\! \times \!b$ matrix, each row corresponds to an independent hash function, which maps $n$-dimensional to $b$-dimensional space~\cite{li2019privacy}, $b\!<\!<\!n$, where 
		\begin{gather} 
			\pmb{g}_i=\begin{bmatrix} {g}_{i,1},{g}_{i,2}, {g}_{i,3},\dots,{g}_{i,k},\dots,{g}_{i,n} \end{bmatrix}^\mathrm{T},
			\label{g_vector}\\
			{h}_1 \dots h_{a} \colon \{1 \dots n\} \rightarrow \{1 \dots b\}
			\label{hash}
		\end{gather}
		An overview of sketch mechanism is shown in Fig.~\ref{fig:sketch1}, including compression and decompression. When client $i$ compresses the gradient vector $\pmb{g}_i$, sketch model $S_i$ uses $a$ independent hash functions to map the $\pmb{g}_i$ to different columns within $a$ rows. Intuitively, for ${g}_{i,k}$, the column position $v$ in the $u$-th row is:
		\begin{equation} 
			\label{get_v}
			v=h_{u}(k),
		\end{equation}
		where $1\le u\le a,\ 1\le v\le b,\ 1\le k\le n$. So the compression position of ${g}_{i,k}$ in sketch model is $S_i[u][v]$. Then the ${g}_{i,k}$ is accumulated into $S_i[u][v]$. Due to $b$ being much smaller than $n$, there may be multiple gradients accumulated at the same position. For the position $S_i[u][v]$, the gradients accumulated at it are $G_{u,v}=\{g_{i,k}|h_u(k)=v\}$. 
		\par The traditional sketch mechanism fixes the size, clients with poor bandwidth experience longer times when uploading the sketch models. Additionally, multiple gradients accumulated at the same position result in poor accuracy when decompressing the sketch model to recover the global model.
		\par To achieve adaptive bandwidth compression and improve accuracy in local model, unlike the general sketch mechanism, we fix columns and elastically adjust its rows, using the coefficient of variation to filter values that are mapped to the same position. 
		
		
		 \par Suppose the sketch model $S_i$ of client $i$ is an $a_i \!\times \! b$ matrix, where rows $a_{i}$ are obtained based on data volume $D_{i}$ and columns $b$:
		\begin{equation} 
			\label{ai}
			a_i=\lfloor \frac{D_i}{b} \rfloor.
		\end{equation}
		In compression operator $\mathcal{C(\cdot)}$, when $\pmb{g}_i$ complete the mappings, we process the values of each position in sketch. For the position $S_i[u][v]$, we process its values $G_{u,v}$:
		\begin{equation}
			\setlength\abovedisplayskip{6pt}
			\setlength\belowdisplayskip{6pt}
			\label{deqn_ex9}
			S_i[u][v]=\begin{cases}
				{mean}(G_{u,v}),&{\text{if}}\ \eta \leq 0.5, \\
				\text{max}\{G_{u,v}\}, &{\text{otherwise.}}
			\end{cases}
		\end{equation} 
		where 
		\begin{gather} 
			mean(G_{u,v})=\frac{1}{|\!|G_{u,v}|\!|} \sum_{g_{i,k}\in G_{u,v}} g_{i,k},
			\label{mean}
		\end{gather}
		$\eta$ is the coefficient of variation of $G_{u,v}$. Thus, client $i$ obtains compression model $S_i$, and then uploads it to the server, where the data volume of upload is $D'_i=a_i\times b$.

		\par In aggregation operator $\mathcal{AGG(\cdot)}$, due to each client has different $b_i$ and $D_i$, the $a_i$ is differ. As a result, the server can not aggregate sketch models.
		In this regard, we have two steps: first, size alignment of sketch models. For all sketch models from clients, the maximum number of rows $a_{max}$ is:
		\begin{equation}
			a_{max} = \max_{1\le i\le C} \{a_i\},
			\label{amax}
		\end{equation}
		when $a_i \neq a_{max}$, we fill the sketch model $S_i$ with zeros to $a_{max}$ rows, resulting in $\widetilde{S}_i$:
		\begin{equation}
			\widetilde{S}_i=\begin{array}{c@{\hspace{-5pt}}l}   
					\begin{bmatrix}  
					*&* &* &\cdots&*\\  
					\vdots & \vdots & \vdots &\ddots & \vdots \\  
					* & * & * & \cdots & *\\
					0      & 0      & 0     &\cdots & 0\\
					\vdots & \vdots & \vdots &\ddots & \vdots \\
					0      & 0      & 0    &\cdots & 0\\
					   
				\end{bmatrix}
				&\begin{array}{l}\left.\rule{0mm}{7mm}\right\}a_i\\   
					\\\left.\rule{0mm}{7mm}\right\}a_{max}\!-\!a_i    
				\end{array}\\[-4pt] 
				\begin{array}{c}
					\underbrace{\rule[0mm]{25mm}{0mm}}_b\end{array}& 
			\end{array}
		\end{equation}
		\par Second, aggregate all sketch models by calculate row-wise averages. To begin with, we obtain $A$, which is the number of non-zero in each row of all sketch models. 
		Then we aggregate the sketch models:
		\begin{gather}
			S_{agg}[u]=\frac{\sum\limits_{i=0}^{C} \widetilde{S}_i[u]}{A[u]},
			\label{avg}
		\end{gather}
		where $S_{agg}$ is the aggregated sketch model.

		Next, each client performs decompress operator $\mathcal{D(\cdot)}$ after downloading the updated sketch model $S_{agg}$ from the server:
		\begin{equation}
			\label{median}
			\overline{g}_k =\text{Median}\{S_{agg}[u][h_{u}(k)]: 1\le u \le a_{max}\}.
		\end{equation}

		\begin{figure}[t]
			
			\centering
			\includegraphics[scale=0.4]{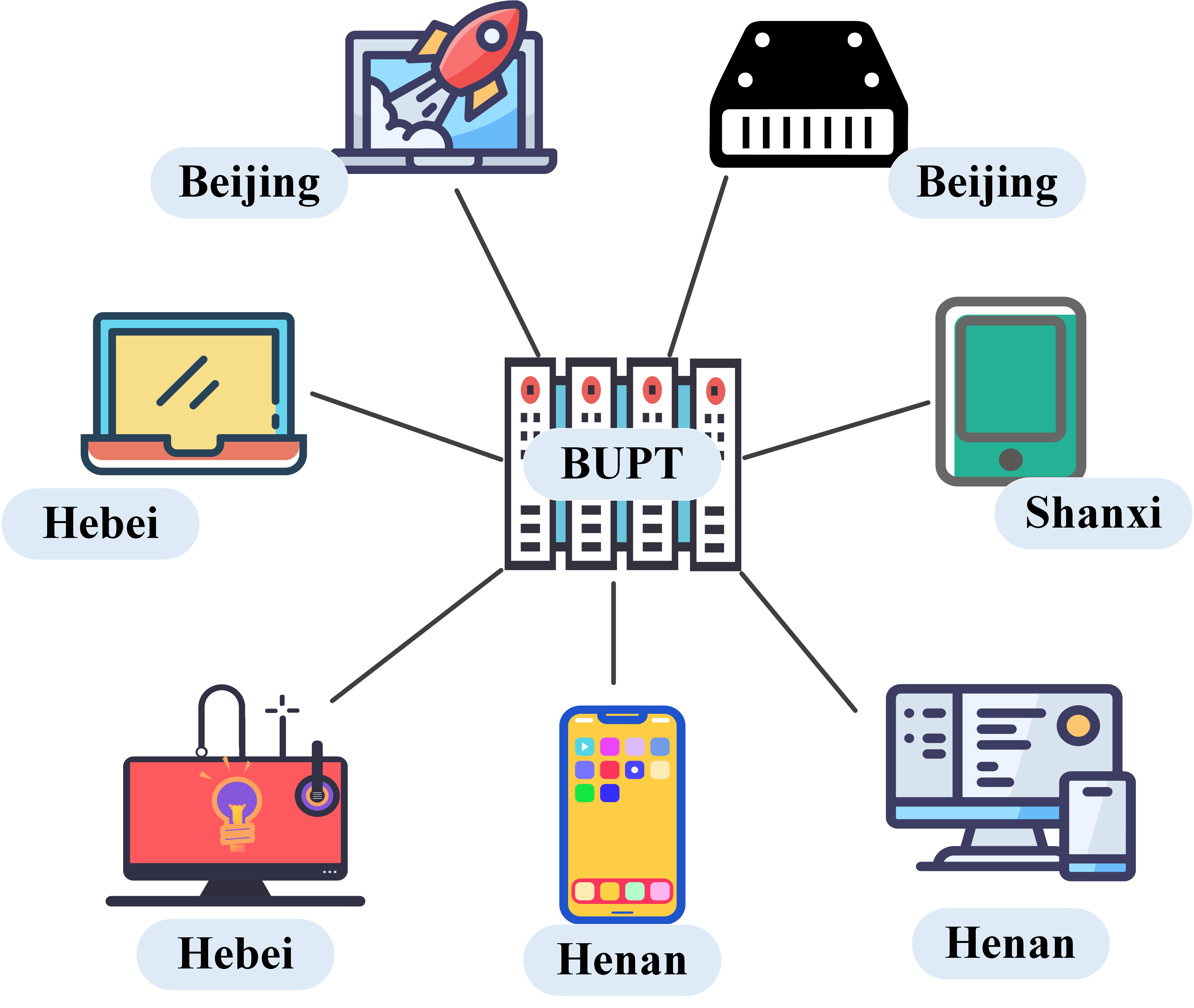}
			\caption{Network topology of bandwidth collection.}
			\label{fig:tuopu}
		\end{figure}

		\begin{table}[ht]
			\centering
			\captionsetup[table]{position=above}
			\caption{models and parameters in experiments.}
			\renewcommand{\arraystretch}{1.1}
			\setlength\tabcolsep{1.1 pt}
			\label{params}
			\begin{tabular}{cccccc}
				
				\toprule [1.5pt]
				
				\multirow{2}{*}{\textbf{Dataset}} & \multirow{2}{*}{\textbf{Algorithm}} & \multirow{2}{*}{\textbf{Model}} &  \multirow{2}{*}{\textbf{Learn Rate}} & \multicolumn{2}{c}{\textbf{Sketch}} \\ 
				&&& & \textbf{Row} & \textbf{Column} \\

				\midrule 
				\multirow{3}{*}{FEMNIST} & FedAvg~\cite{mcmahan2017communication} & ResNet~56 & 0.001 & - & -  \\
				\multirow{3}{*}{}      & SketchFL~\cite{liu2019enhancing} & ResNet~56 & 0.001 & 7  & 60000 \\
				\multirow{3}{*}{}     & AdapComFL & ResNet~56 & 0.001 & [3,10] & 60000 \\[0.5pt]
				
				\cline{2-6}\rule{0pt}{10pt}
				\multirow{3}{*}{FashionMNIST} & FedAvg~\cite{mcmahan2017communication} & ResNet~44 & 0.002 & - & -  \\
				\multirow{3}{*}{}      & SketchFL~\cite{liu2019enhancing} & ResNet~44 & 0.002 & 7  & 50000 \\
				\multirow{3}{*}{}     & AdapComFL & ResNet~44 & 0.002 & [3,10] & 50000 \\
				\bottomrule [1.5pt] \rule{0pt}{8pt}   
			\end{tabular}
		\end{table}
		
		\begin{figure}[t]
			
			\centering
			\includegraphics[scale=0.55]{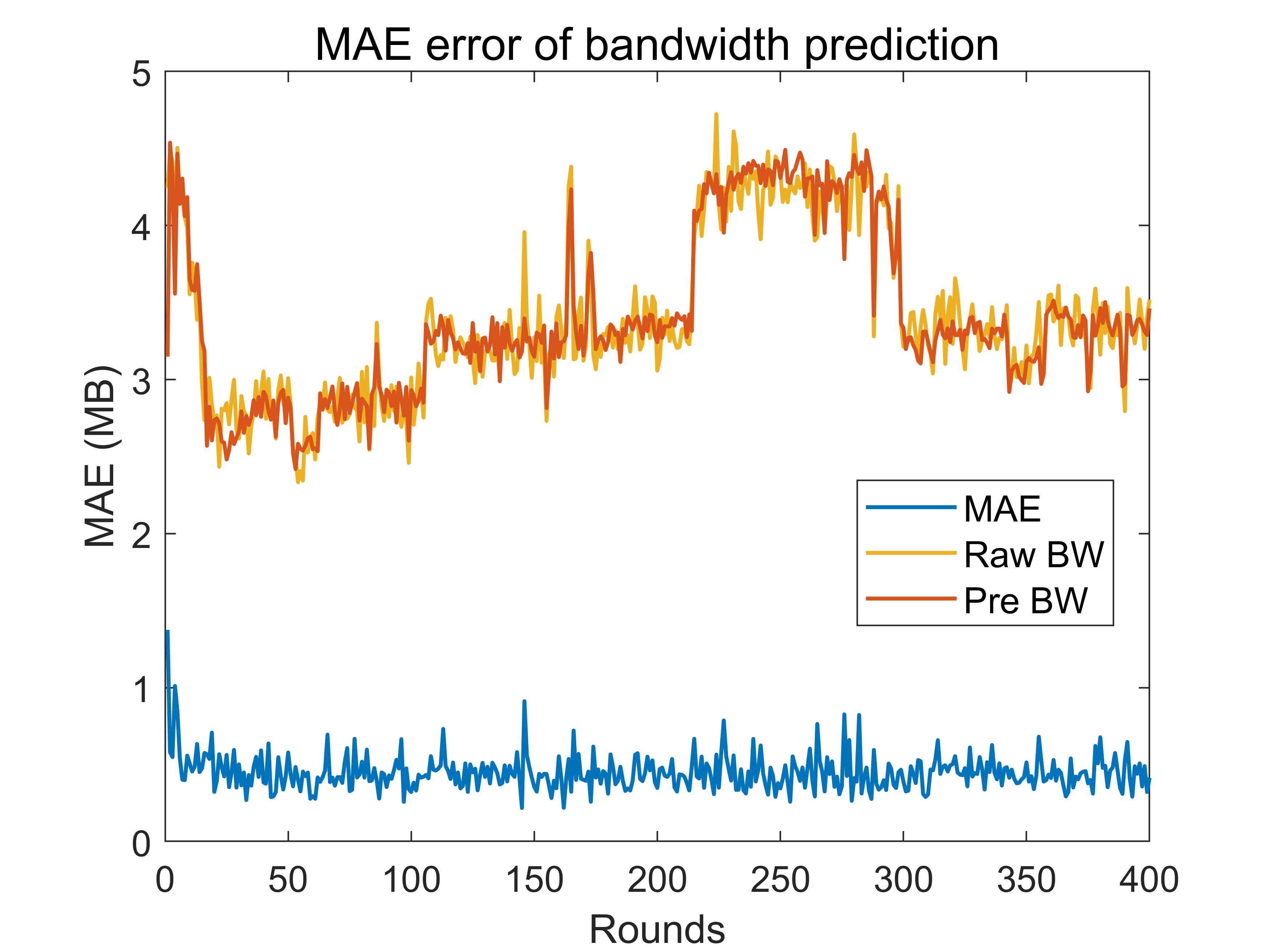}
			\caption{Accuracy results of bandwidth prediction.}
			\label{fig:mae}
		\end{figure}

		\section{Evaluation}
		In this section, we present the setups and results of the experiment. The setups mainly have five parts: environments, datasets, models, parameters, and metrics. In the results, we evaluate the accuracy of bandwidth prediction, then show the performance of the AdapComFL algorithm, and compare it with other algorithms.
		\subsection{Setups}
		\par {\bf{Environment: }} In order to simulate federated learning tasks, the experiments are implemented based on PyTorch~\cite{paszke2019pytorch} with an NVIDIA GeForce RTX 3090 GPU with 24 GB of memory. 
		
		
		\par {\bf{Datasets: }}To evaluate the bandwidth prediction and the AdapComFL algorithm, we collect real bandwidth datasets and use two benchmark datasets.
		\par For bandwidth datasets, we build a network topology through Iperf~\cite{tirumala1999iperf} to collect bandwidth data which is show in Fig.~\ref{fig:tuopu}. The collection equipments are different types of personal computers (PC), cell phones, and mini-PC, the locations are Beijing, Hebei, Henan, and Shanxi. We set up a server as the central node at the Beijing University of Posts and Telecommunications (BUPT). The 7~peripheral nodes have two serves: they function as probe nodes for collecting bandwidth data; they act as client nodes in federated learning. From July 1, 2022, to July 30, 2022, the network topology collects bandwidth continuously by seconds and every 24 hours generating one group. Thus, each probe node has a bandwidth dataset with 30 groups. Additionally, In upload process, the maximum time $\text{T}$ for uplink communication in Eq.~(\ref{deqn_ex6}) is 0.5~s. 

		\par For benchmark datasets, we chose the public datasets FEMNIST \cite{caldas2018leaf} and FashionMNIST \cite{xiao2017fashion}, which are processed to meet non-iid. 
		FEMNIST is a handwritten digit recognition dataset containing 62 categories. FashionMNIST is an image classification dataset of fashion products for 10 categories.

		\begin{figure*}[ht]
			\centering
			\subfloat[]{\includegraphics[scale=0.4]{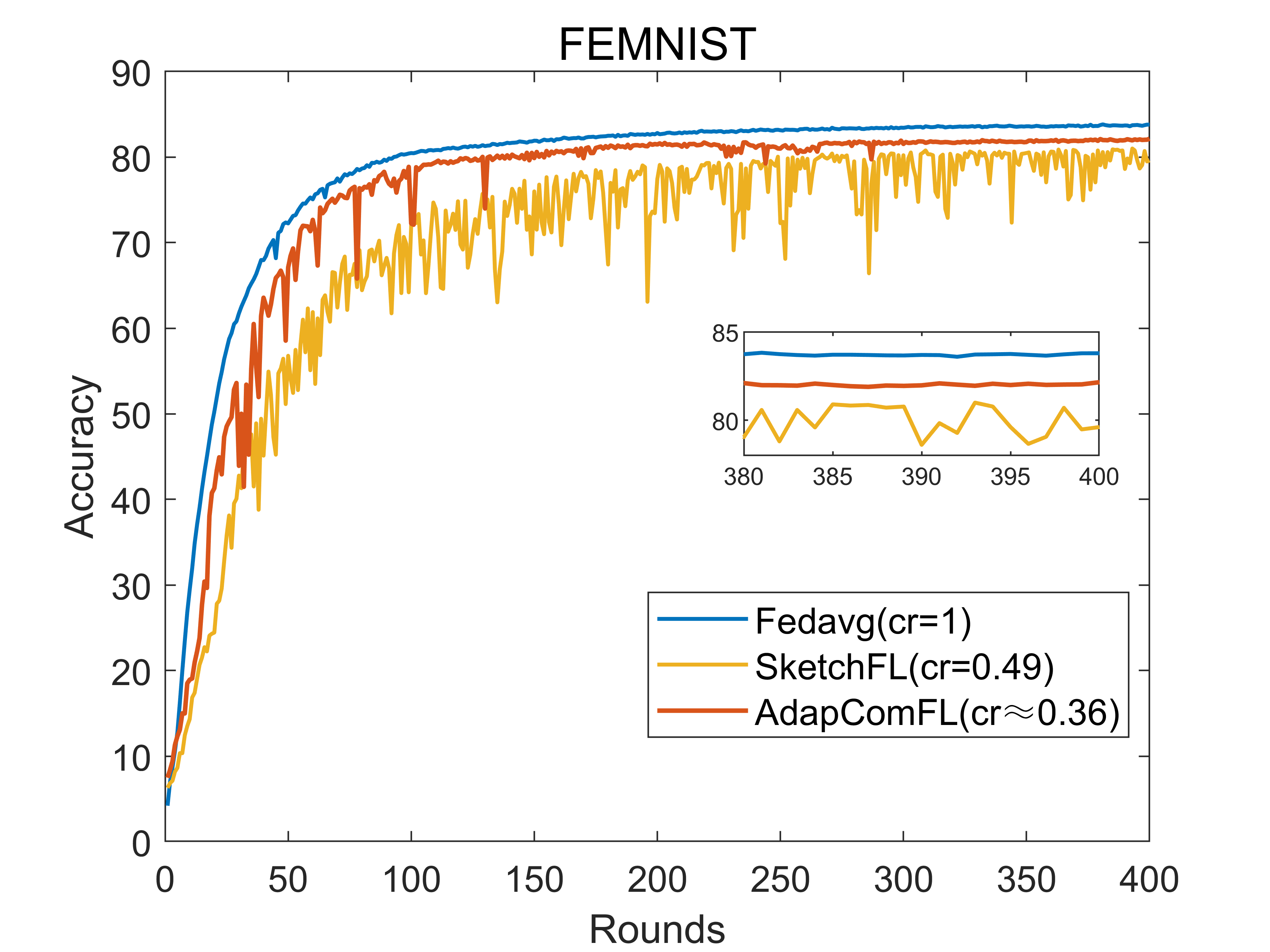}%
				\label{femnist_acc}}
			\hfil
			\subfloat[]{\includegraphics[scale=0.4]{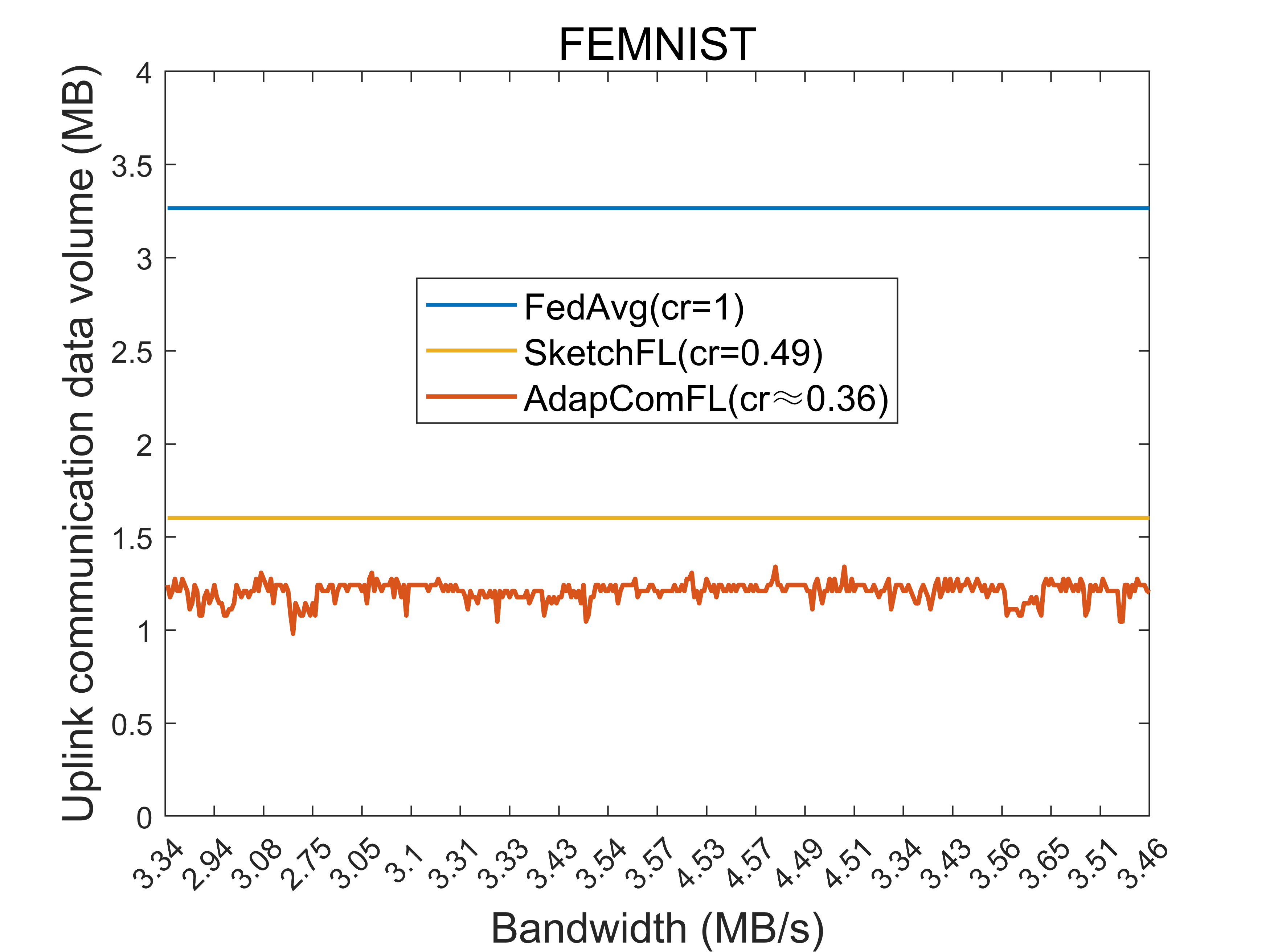}%
				\label{femnist_cost}}
			\hfil
			\subfloat[]{\includegraphics[scale=0.4]{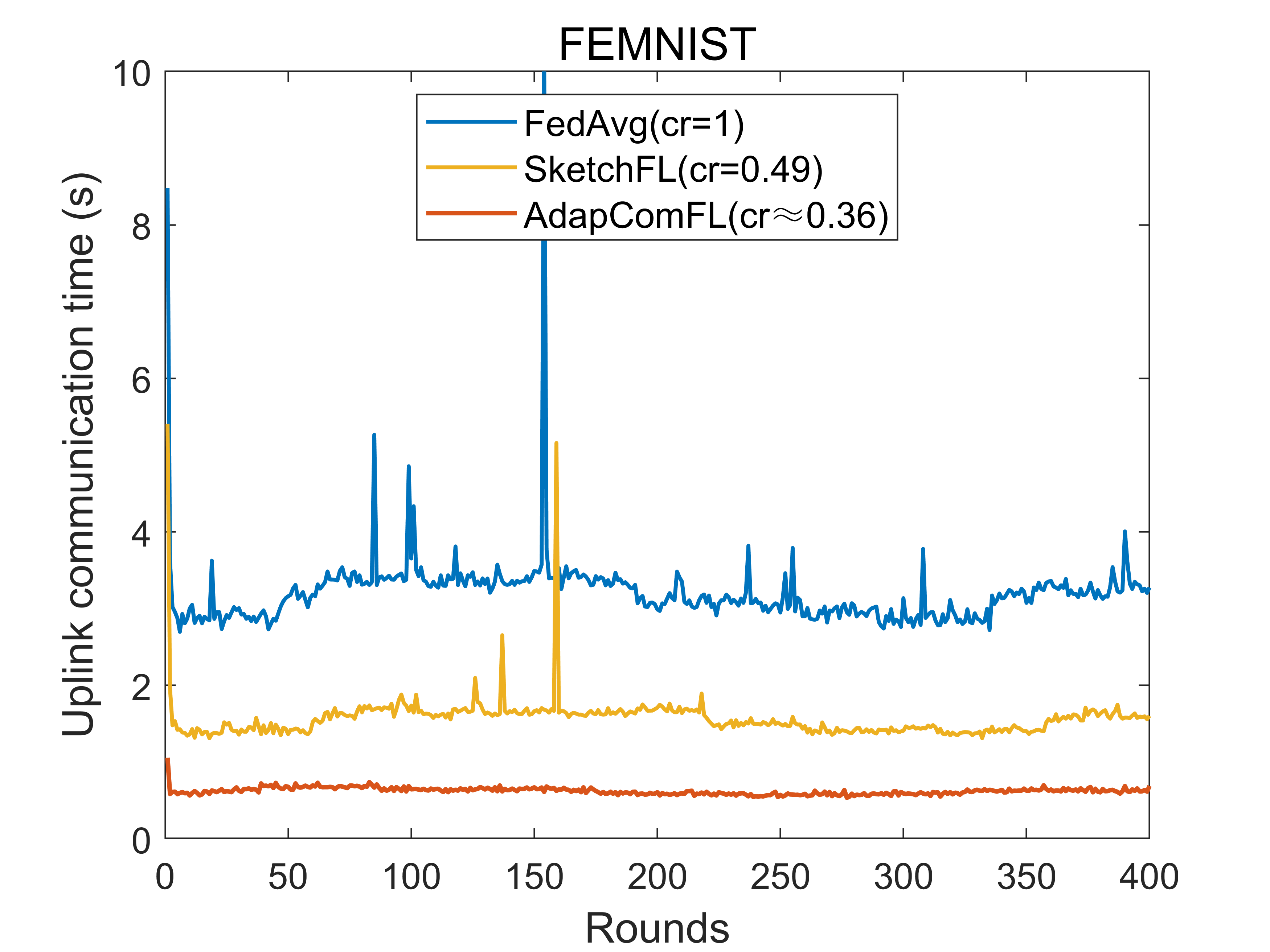}%
				\label{femnist_time}}
			\caption{Results of the FedAvg, SketchFL and AdapComFL (ours) on FEMNIST. (a) The accuracy among FedAvg, SketchFL, and AdapComFL, reflects the performance of each algorithm. (b) The data volume of uplink communication among FedAvg, SketchFL and AdapComFL. (c) The time of uplink communication among FedAvg, SketchFL and AdapComFL. }
			\label{fig:femnist}
		\end{figure*}
		\begin{figure*}[ht]
			\centering
			\subfloat[]{\includegraphics[scale=0.4]{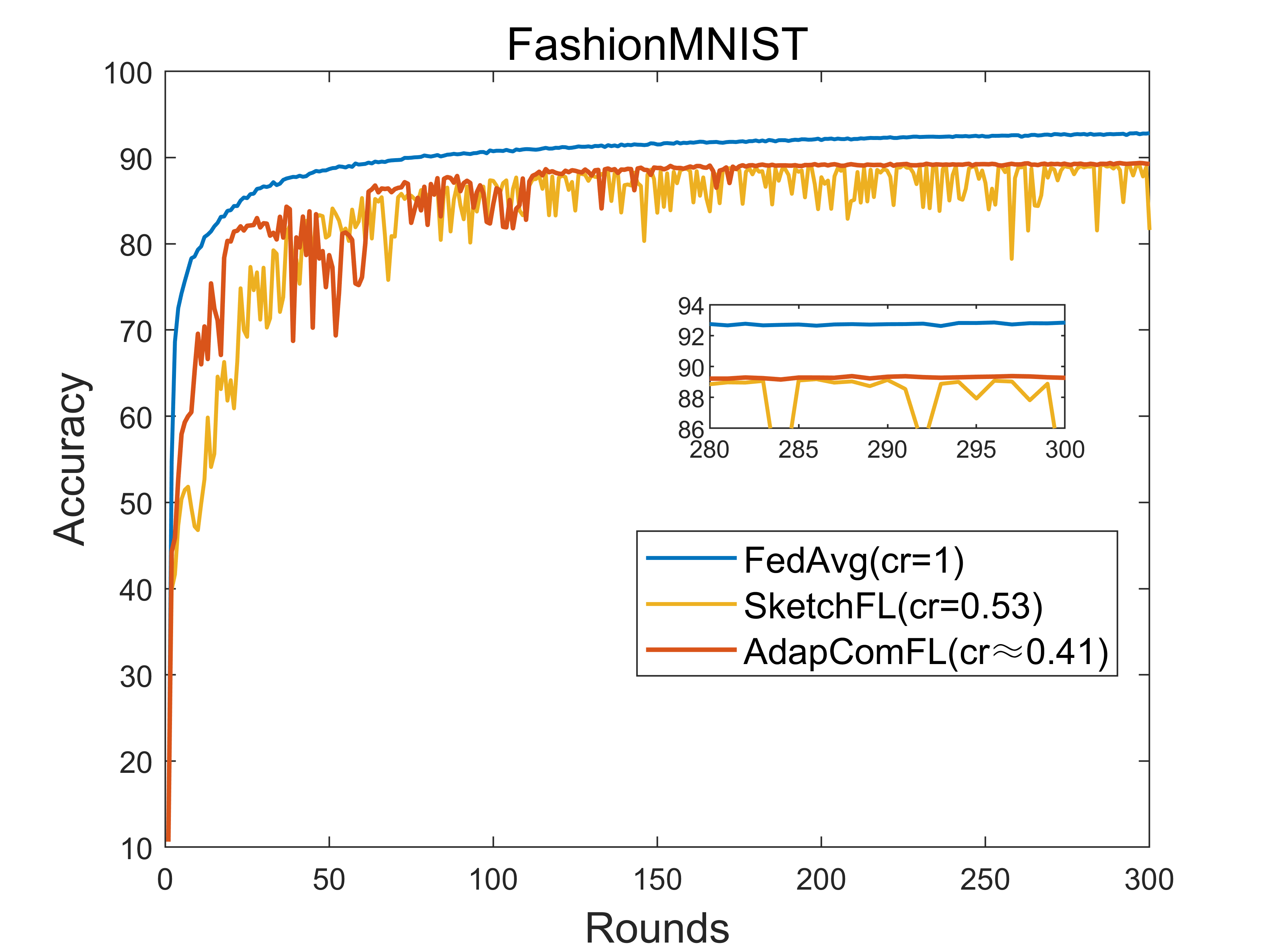}%
				\label{fashion_acc}}
			\hfil
			\subfloat[]{\includegraphics[scale=0.4]{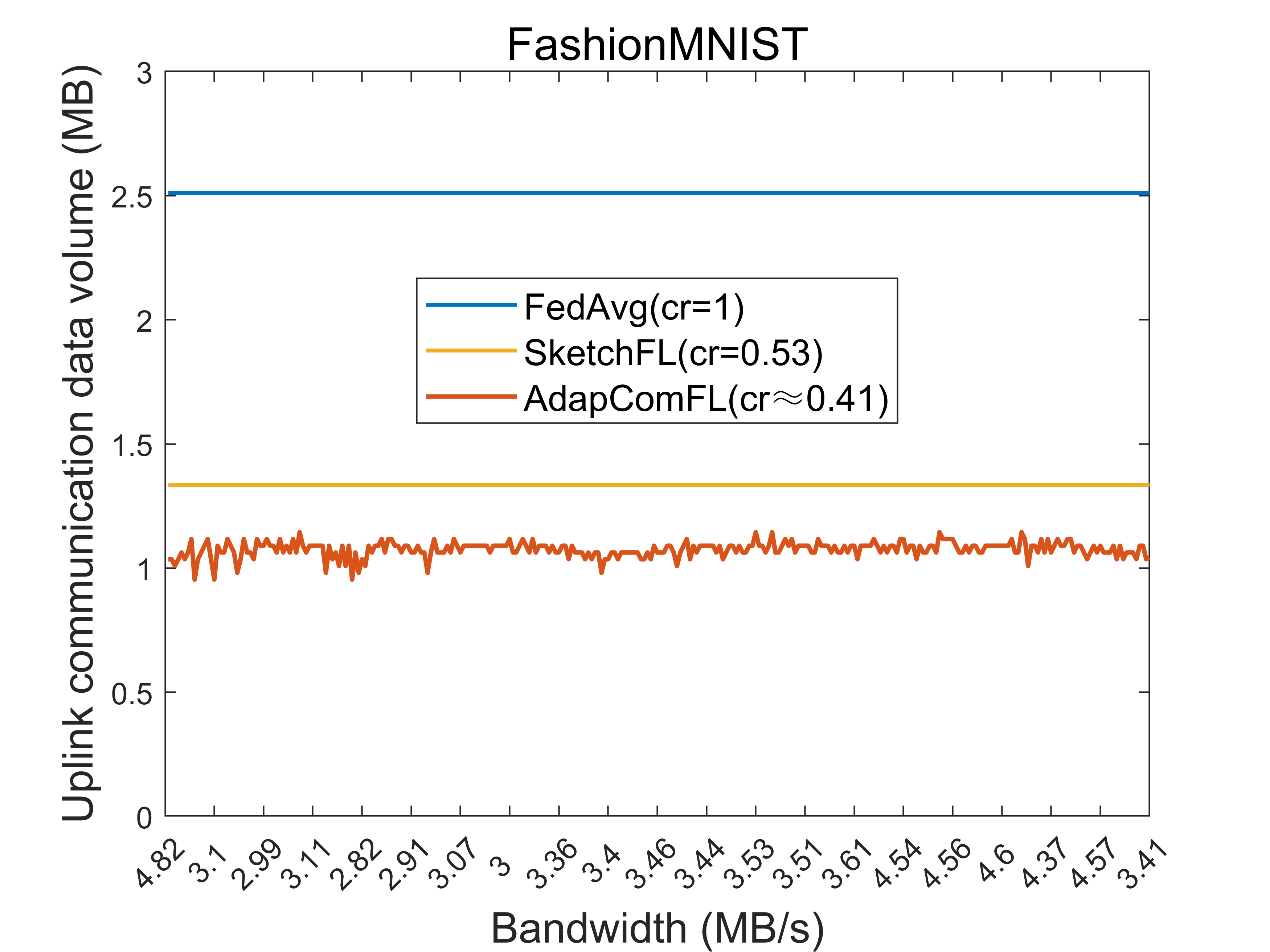}%
				\label{fashion_cost}}
			\hfil
			\subfloat[]{\includegraphics[scale=0.4]{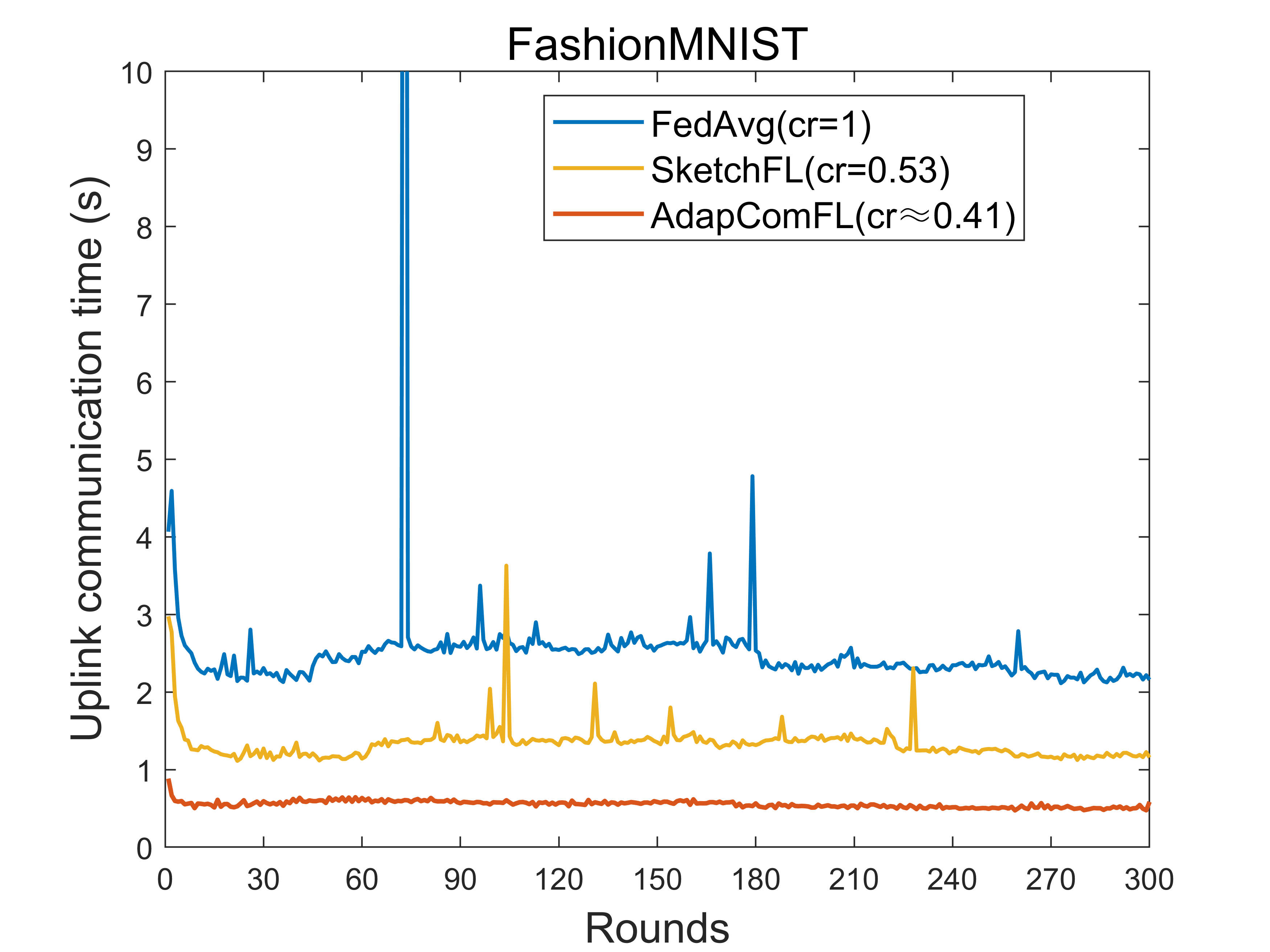}%
				\label{fashion_time}}
			\caption{Results of the FedAvg, SketchFL and AdapComFL (ours) on FashionMNIST. (a) The accuracy among FedAvg, SketchFL, and AdapComFL, reflects the performance of each algorithm. (b) The data volume of uplink communication among FedAvg, SketchFL and AdapComFL. (c) The time of uplink communication among FedAvg, SketchFL and AdapComFL.}
			\label{fig:fashion}
		\end{figure*}
		
		\par {\bf{Models and Parameters: }}This part describes neural network models and parameters in experiments. In the LSTM network used in bandwidth prediction, the input layer with 6~dimensions, the two hidden layers, each with 256 and 128~LSTM units respectively, and the size of the output layer is~1. The algorithms are experimented via Residual Network (ResNet)~\cite{he2016deep}, the models and parameters are shown in Table~\ref{params}. 
		
		

		\begin{table}[t]
			\centering
			\captionsetup[table]{position=above}
			\caption{Comparison table of predicted bandwidth (Pre BW) and real bandwidth (Raw BW) for each client.}
			\label{tab:bandwidth}
			\setlength\tabcolsep{6.5 pt}
			\begin{tabular}{cccccccc}
				
				\toprule [1.5pt]
				\multirow{2}{*}{\textbf{ }} & \multicolumn{7}{c}{\textbf{Clients}} \\ 
				& \textbf{\#0} & \textbf{\#1} & \textbf{\#2} & \textbf{\#3} & \textbf{\#4} & \textbf{\#5} & \textbf{\#6} \\
				\midrule 
				Raw BW & 12.72 & 4.81 & 2.59 & 1.73 & 0.42 & 1.78 & 0.93  \\
				Pre BW & 12.82 & 4.76 & 2.51 & 1.77 & 0.41 & 1.77 & 0.94 \\ 
				\bottomrule [1.5pt]     
			\end{tabular}
		\end{table}		
				
		\begin{figure}[htbp]
			
			\centering
			\includegraphics[scale=0.55]{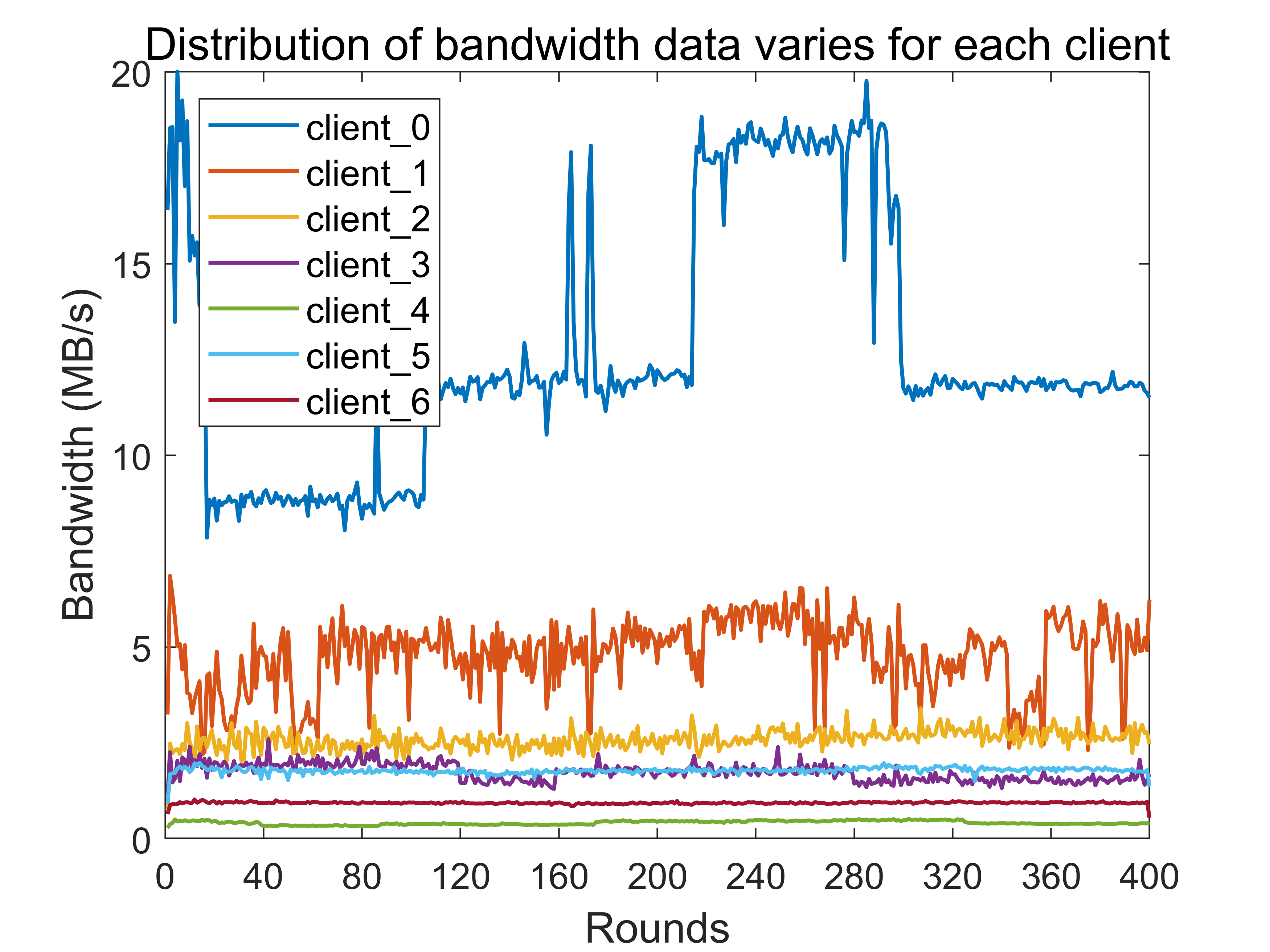}
			\caption{Distribution results of bandwidth data among clients.}
			\label{fig:total_bw_distribution}
		\end{figure}
		\par {\bf{Metrics: }}In order to evaluate accuracy and communication efficiency of AdapComFL algorithm, some metrics are selected.
		\par In accuracy, we consider the bandwidth prediction and the algorithm. For bandwidth prediction, we evaluate the accuracy by Mean Absolute Error (MAE). 
		The formula of $M\!A\!E$ is given by
		\begin{equation}
			\label{deqn_ex18}
			M\!A\!E= \frac{1}{p}\sum_{q=1}^{{p}} 
			|{{{b'}}_{q}}-{{b}_{q}}|,
		\end{equation}
		where $p$ is the number of $b$. 
		The accuracy of prediction increases as $M\!A\!E$ decreases. 
		The accuracy of alogorithm is:
		\begin{equation}
			\label{acc}
			acc=\frac{N_{t}}{N}\times 100\%,
		\end{equation}
		where $N_{t}$ is the number of correctly classified samples. 
		\par In communication efficiency, we consider the compression ratio $cr\!=\!\frac{D'_{}}{D_{avg}}$. 
		where ${D_{avg}}$ is the data volume of uplink communication by clients in the FedAvg algorithm, which is a constant, and $cr$ reflects the compression degree of the model, which as $cr$ decreases, the compression level increases, the time $T'$ and the data volume $D'$ of uplink communication are reduce.
		The compression ratio

		\begin{figure*}[ht]
			\centering
			\subfloat[]{\includegraphics[scale=0.4]{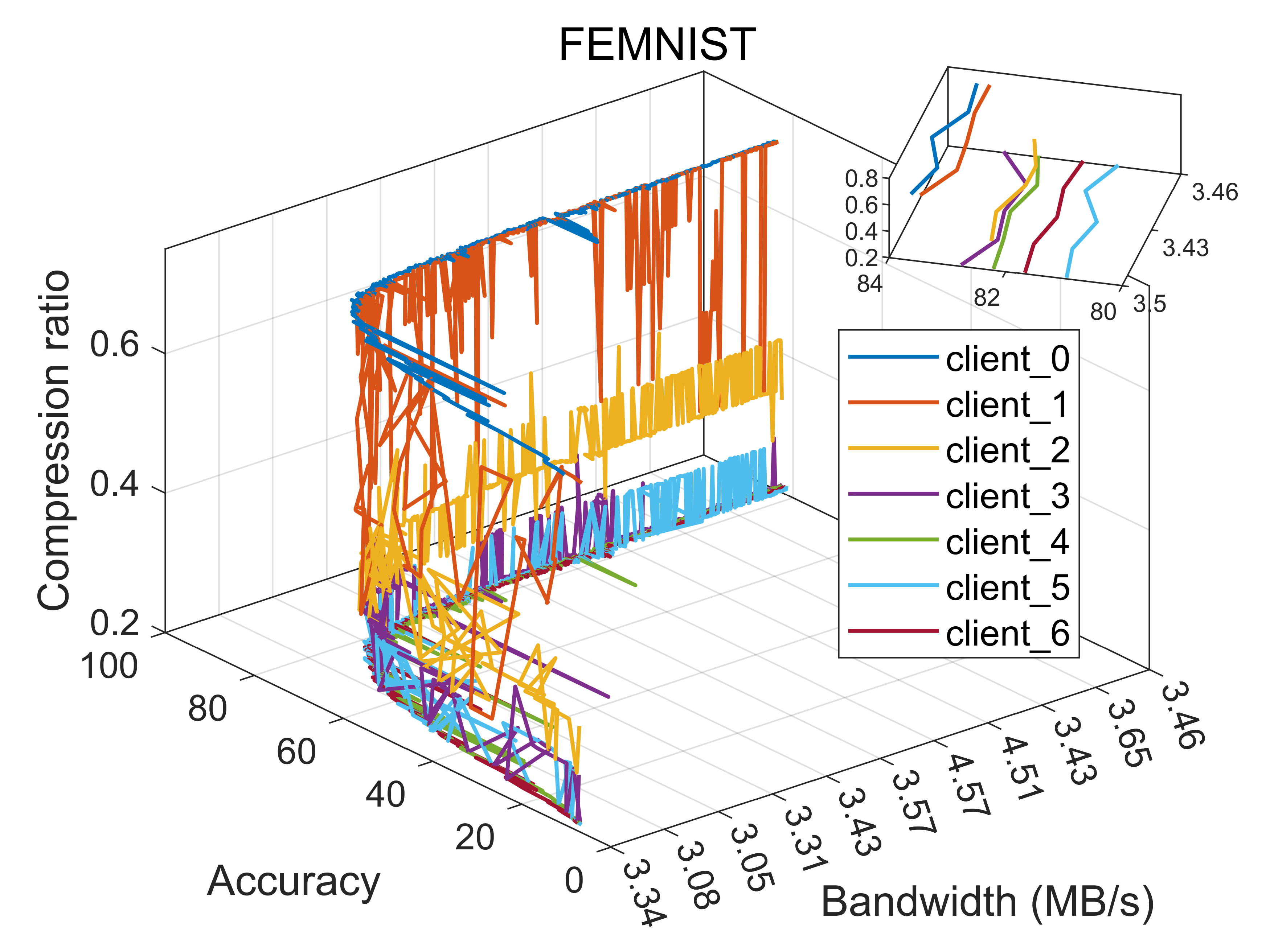}%
				\label{femnist_user_3d}}
			\hfil
			\subfloat[]{\includegraphics[scale=0.4]{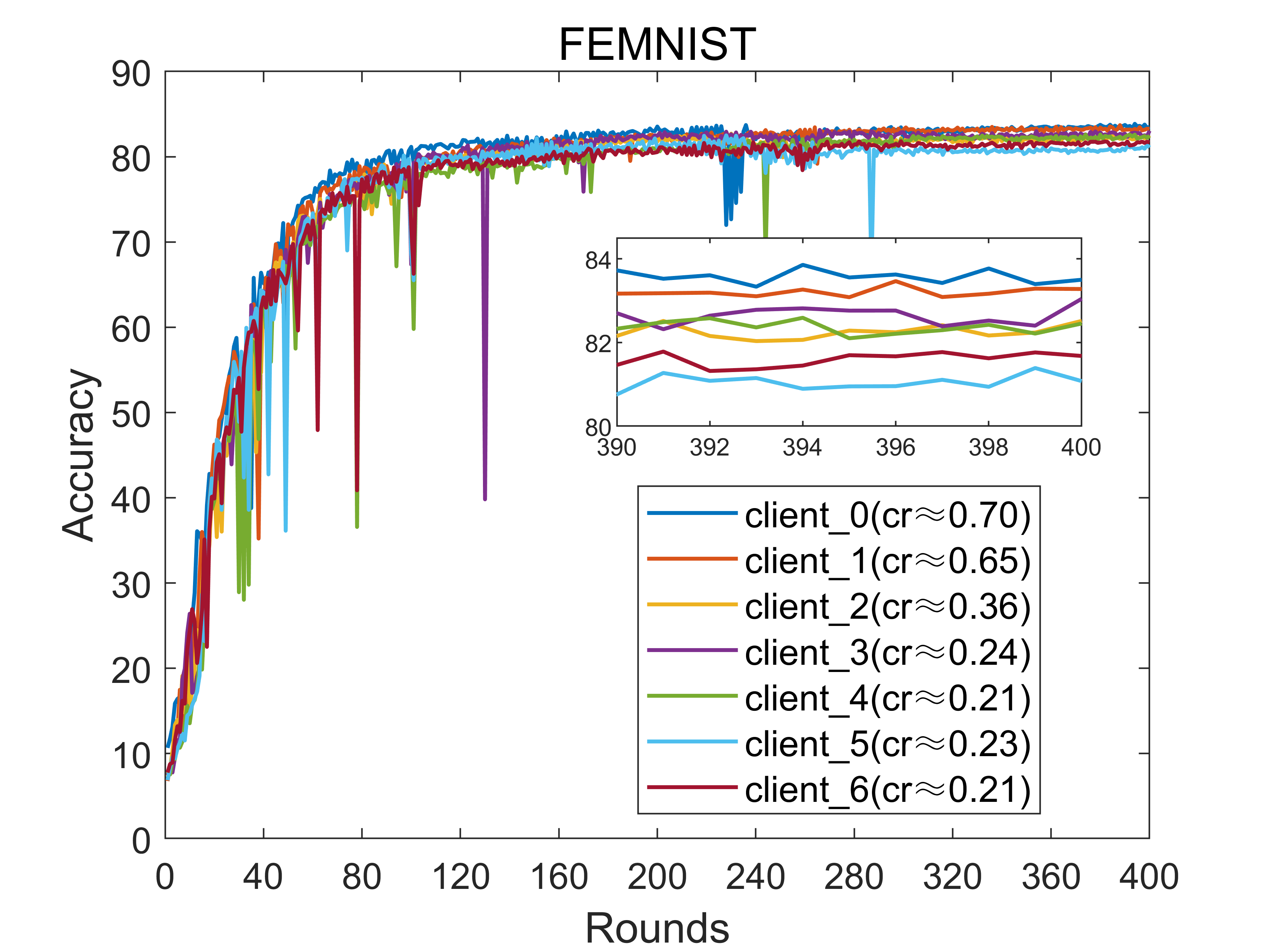}%
				\label{femnist_client_acc}}
			\hfil
			\subfloat[]{\includegraphics[scale=0.4]{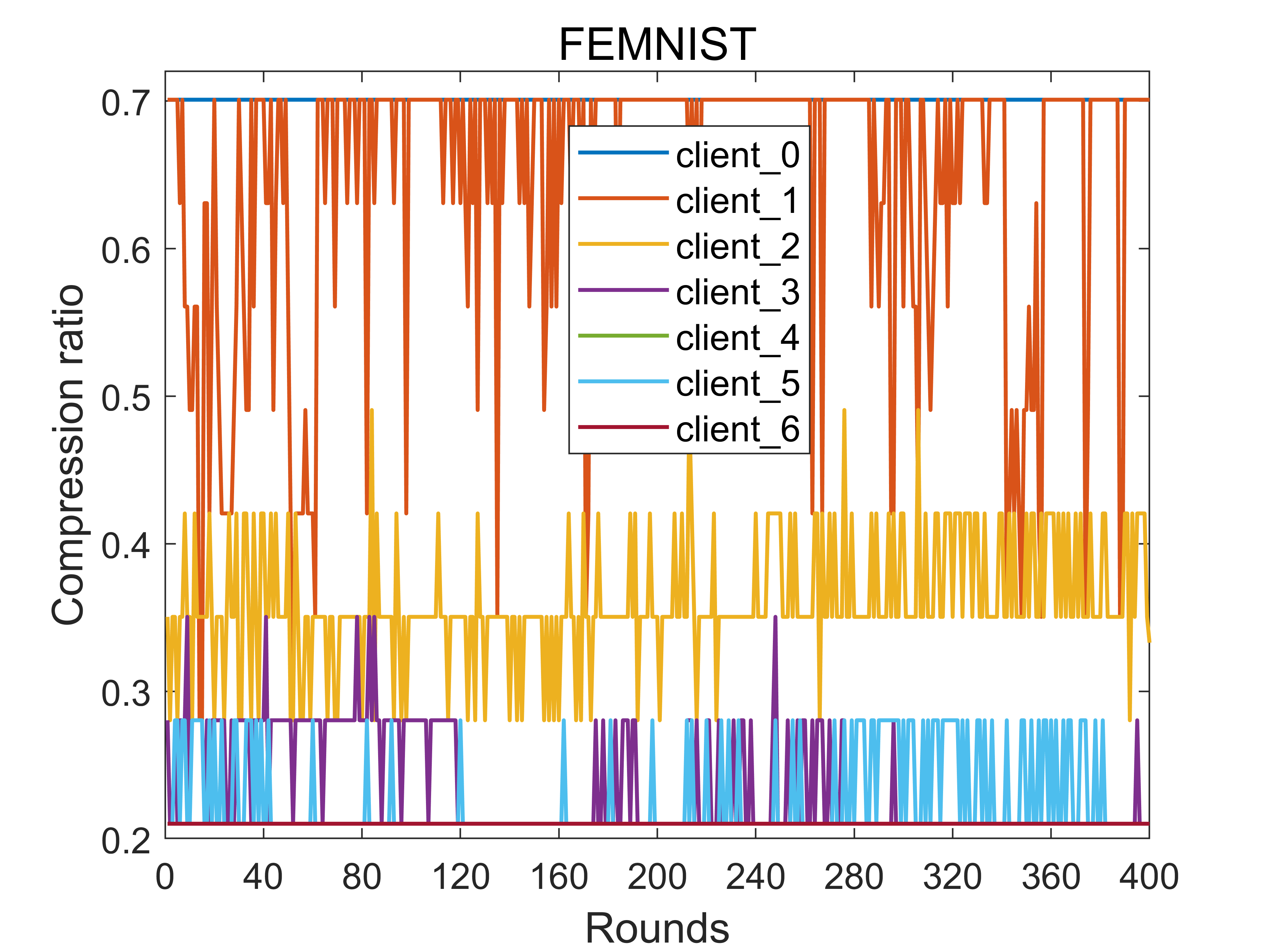}%
				\label{femnist_client_ratio}}
			\caption{Results of each client using AdapComFL (ours) on FEMNIST. (a) The figure about bandwidth, accuracy, and compression ratio, which reflects the performance and the level of model compression for each client. (b) The accuracy shows the performance of the AdapComFL algorithm on each client. (c) The compression ratio illustrates the level of model compression for each client. }
			\label{fig:client_femnist}
		\end{figure*}
		
		\begin{figure*}[ht]
			\centering
			\subfloat[]{\includegraphics[scale=0.4]{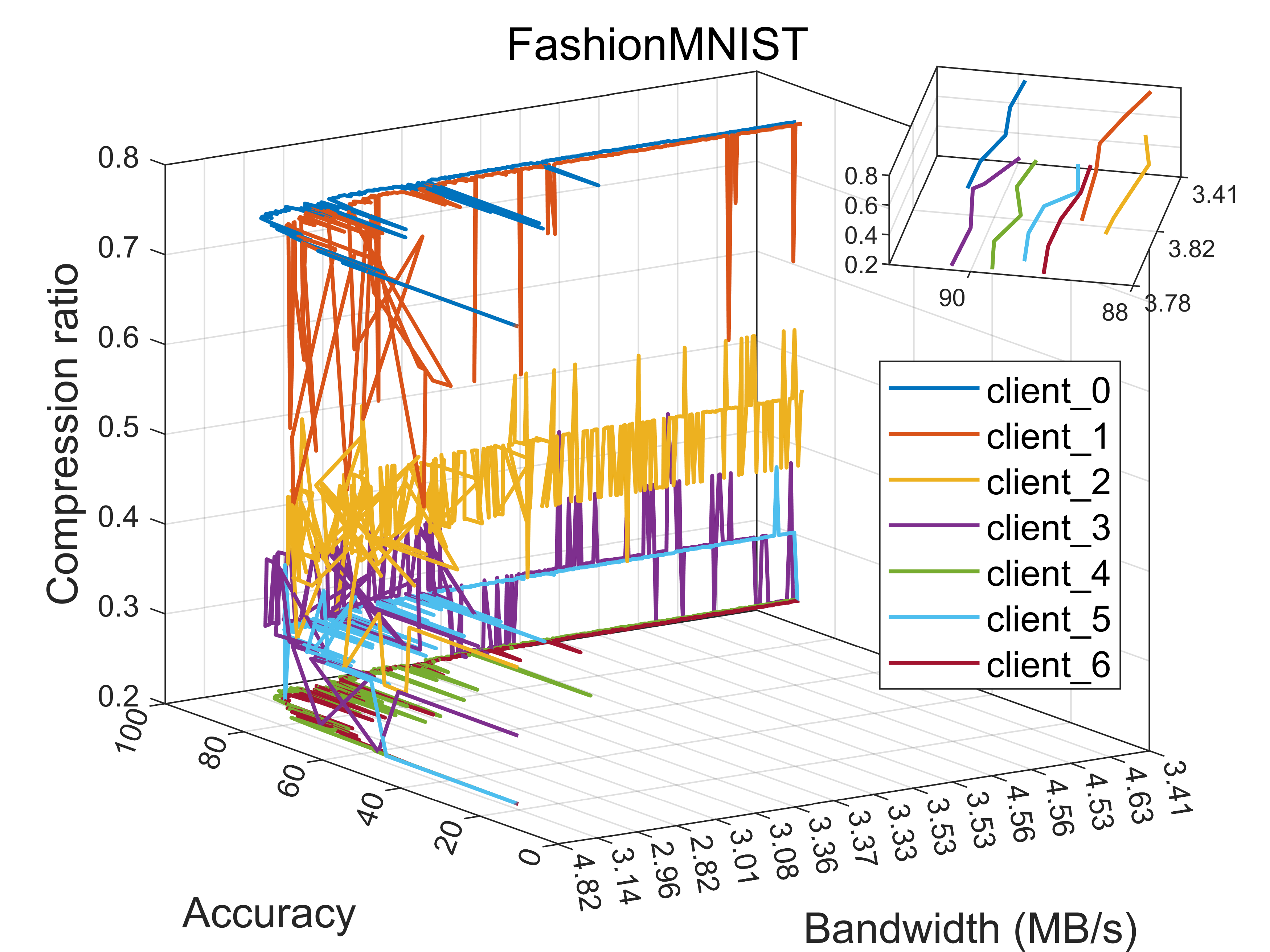}%
				\label{fashion_user_3d}}
			\hfil
			\subfloat[]{\includegraphics[scale=0.4]{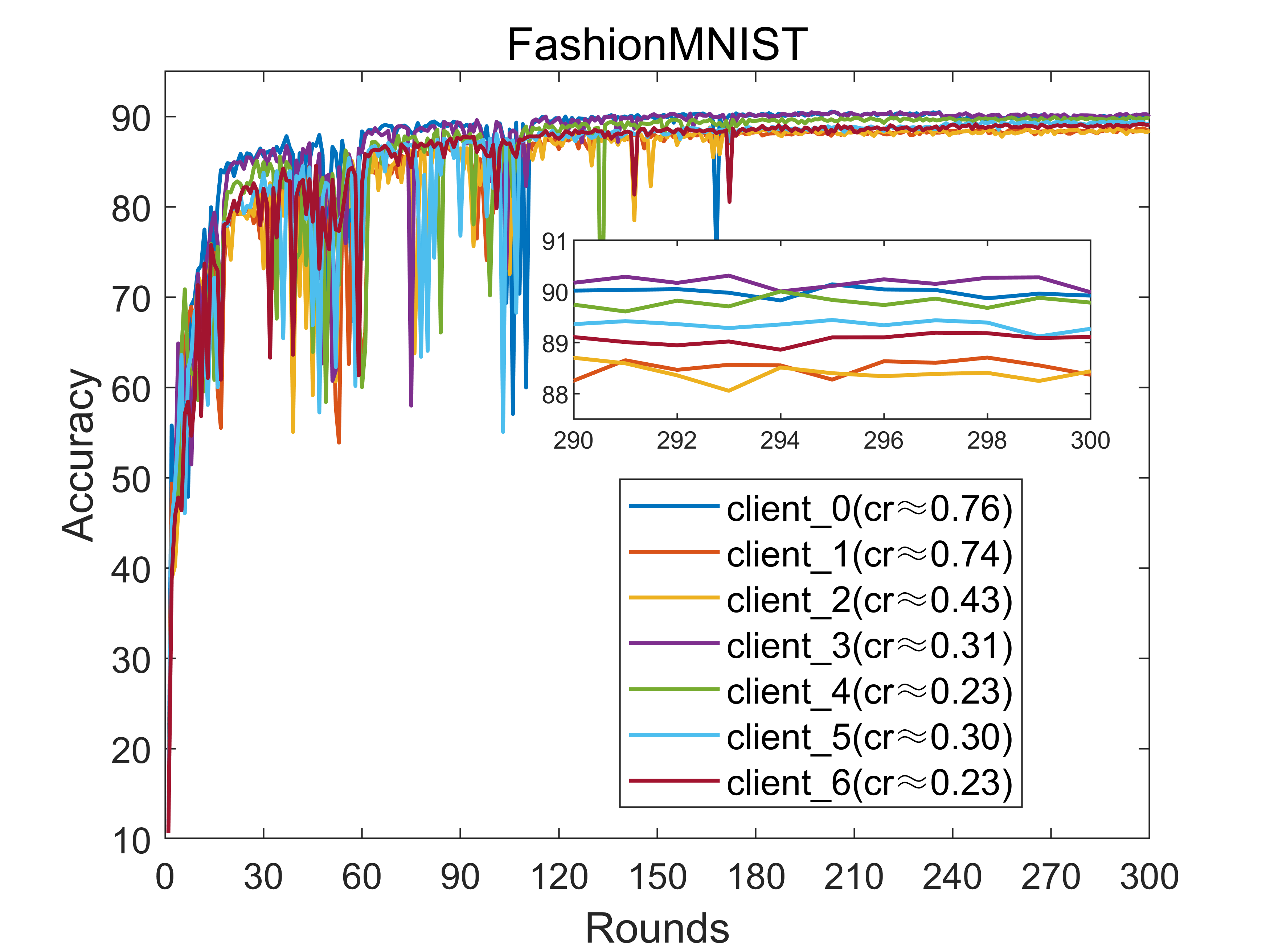}%
				\label{fashion_client_acc}}
			\hfil
			\subfloat[]{\includegraphics[scale=0.4]{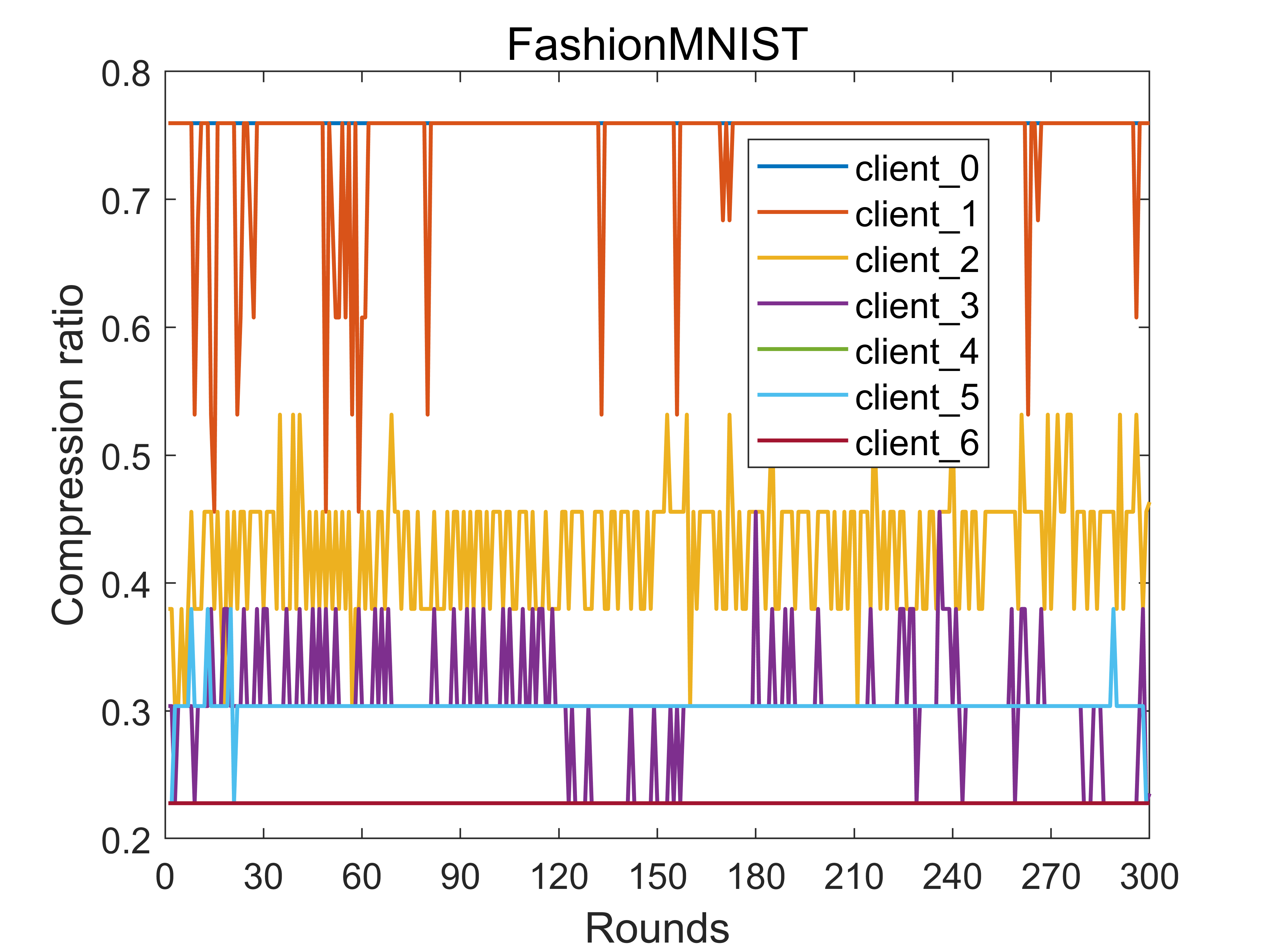}%
				\label{fashion_client_ratio}}
			\caption{Results of each client using AdapComFL (ours) on FashionMNIST. (a) The figure about bandwidth, accuracy, and compression ratio, which reflects the performance and the level of model compression for each client. (b) The accuracy shows the performance of the AdapComFL algorithm on each client. (c) The compression ratio illustrates the level of model compression for each client. }
			\label{fig:client_fashion}
		\end{figure*}
		

		\subsection{Results}
		\par In this section, we compare the AdapComFL algorithm with other algorithms for accuracy and communication efficiency. 

		\par {\bf{Accuracy: }}We show the accuracy of bandwidth prediction and algorithms.
		\par The accuracy of bandwidth prediction can be shown in Fig.~\ref{fig:mae}. It illustrates the accuracy of bandwidth prediction through overall error, where Raw BW (MB) is the average of bandwidth real data, Pre BW (MB) is the average of bandwidth prediction data. From Fig.~\ref{fig:mae}, the Pre BW fits the Raw BW by LSTM prediction and the error is around 0.5 MB, which implies AdapComFL has favorable accuracy in bandwidth prediction. 
		Table~\ref{tab:bandwidth} shows each client's average value of Raw BW and Pre BW. The similarity of the data further confirms the validity of the bandwidth prediction.
		\par We then present the accuracy of AdapComFL on FEMNIST and FashionMNIST, comparing it with FedAvg and SketchFL, as shown in Fig.~\ref{fig:femnist}\subref{femnist_acc} and \ref{fig:fashion}\subref{fashion_acc}. In these figures, the AdapComFL algorithm at a lower compression ratio outperforms slightly inferior to FedAvg in accuracy but has higher accuracy and more stable results than SketchFL. The compression method of the sketch is lossy, but we maintain favorable accuracy by improving the sketch mechanism.
		The experimental results imply that the improvement is effectual. In particular, it is even more evident in the FEMNIST dataset. The above accuracy focus on the aggregation models. For each client in our AdapComFL, the accuracy is as shown in Fig.~\ref{fig:client_femnist}\subref{femnist_client_acc},~\ref{fig:client_fashion}\subref{fashion_client_acc}. Despite variations in compression ratios, the clients present high-accuracy.
		\par {\bf{Communication efficiency: }}We compared the communication efficiency of AdapComFL algorithm with other algorithms from communication time, communication data volume, and compression ratio. 
		\par The communication data volume is shown in Fig.~\ref{fig:femnist}\subref{femnist_cost} and \ref{fig:fashion}\subref{fashion_cost}. It can be observed that the FedAvg algorithm presents high and constant data volume due to it uploading uncompressed models. The SketchFL algorithm has constant data volume due to the fixed compression ratio. In contrast, our proposed AdapComFL fluctuates data volume due to the adaptive compression model based on bandwidth conditions. 
		\par We then present the communication time in Fig.~\ref{fig:femnist}\subref{femnist_time} and \ref{fig:fashion}\subref{fashion_time}. FedAvg and SketchFL take longer times to upload, but AdapComFL completes the upload around 0.5~s, which meets the maximum time we have set. It is evident that we achieve communication efficiency. Considering some clients with poor bandwidth conditions, the maximum compression may still exceed the maximum time. 
		\par Actually, the above results are the average outcomes for clients and the fluctuation is not large enough. We show the results of each client from the network bandwidth and performance in our AdapComFL. In network bandwidth, the distribution for each client is shown in Fig.~\ref{fig:total_bw_distribution}. The curves of the clients show significant differences because they have various bandwidth conditions. 
		We then discuss the results of bandwidth, accuracy, and compression ratio in Fig.~\ref{fig:client_femnist}\subref{femnist_user_3d} and~\ref{fig:client_fashion}\subref{fashion_user_3d} on FEMNIST and FashionMNIST. In order to provide a more intuitive understanding of the accuracy and compression ratio for each client, we present Fig.~\ref{fig:client_femnist}\subref{femnist_client_acc},~\ref{fig:client_femnist}\subref{femnist_client_ratio} for FEMNIST, and~\ref{fig:client_fashion}\subref{fashion_client_acc},~\ref{fig:client_fashion}\subref{fashion_client_ratio} for FashionMNIST.
		Before each round of upload, clients aware of and predict the network bandwidth, then compress the model adaptively. The clients with poor bandwidth conditions achieve communication efficiency by heavily compressing the local model, while maintaining high accuracy. The clients with strong bandwidth conditions utilize bandwidth resources effectively by lightly compressing the model, which obtains high accuracy. 
		
		\section{Conclusions}

		In this article, we address the communication efficiency problem with federated learning. To overcome the bandwidth issues in communication efficiency, we propose AdapComFL algorithm, in which each client predicts its bandwidth and adaptively compresses the model gradient with the sketch method. In addition, we improve the sketch-based compression mechanism to alleviate the error. 
		We perform experiments through real bandwidth datasets and benchmark datasets, with the bandwidth datasets collected from the network topology  we build. The experiments compare the AdapComFL algorithm with the FedAvg and SketchFL algorithms in accuracy and communication efficiency. The results demonstrate that AdapComFL predicts the bandwidth with low error, and achieves communication efficiency of federated learning. Moreover, the improved sketch mechanism not only boosts the accuracy of AdapComFL but also leads to more stable results.


	\end{CJK*}	
\end{document}